\documentclass[letterpaper, 10 pt, conference]{ieeeconf}  

\IEEEoverridecommandlockouts                    

\usepackage{cite}
\usepackage{pdflscape}
\usepackage{adjustbox}

\usepackage{algorithm}

\usepackage[
]{algpseudocode}

\usepackage{afterpage}

\usepackage{graphicx}

\usepackage[
    labelformat=simple
]{subcaption}

\usepackage{amsmath}

\usepackage{amssymb}

\usepackage{bbm}

\usepackage{mathrsfs}

\usepackage{mathtools}

\usepackage{xfrac}

\usepackage[Gray,squaren,thinqspace,thinspace]{SIunits}

\usepackage{array}

\usepackage{booktabs}

\usepackage{multirow}

\usepackage{tabu}

\usepackage{tabularx}

\usepackage{tabulary}

\usepackage{pifont}

\usepackage{xcolor}
\usepackage{microtype}
\usepackage{hyperref}
\usepackage[normalem]{ulem}

\usepackage{changepage}
\usepackage{datetime}
\usepackage{tabularx}
\usepackage{xcolor}

\newenvironment{base_cover_letter}[3]{
    \begin{titlepage}
    \def\paperTitle{#1}
    \def\journalName{#2}
    \def\authorName{#3}
    \setlength{\parindent}{0pt}
    \setlength{\parskip}{1.5em}
    \fontsize{10pt}{12pt}\selectfont
    \begin{center}
        \LARGE \textbf{Cover Letter}
    \end{center}
    \bigskip
}{
    Sincerely, \\
    \authorName{}
    \end{titlepage}
}

\AtBeginDocument{\colorlet{defaultcolor}{.}}

\title{\LARGE \bf
Investigation of Factorized Optical Flows as Mid-Level Representations \vspace{-0.5em}
}

\author{Hsuan-Kung Yang$^{1}$, Tsu-Ching Hsiao$^{1}$, Ting-Hsuan Liao$^{1}$*, Hsu-Shen Liu$^{1}$*, Li-Yuan Tsao$^{1}$*, \\Tzu-Wen Wang$^{1}$*, Shan-Ya Yang$^{1}$*, Yu-Wen Chen$^{1}$, Huang-Ru Liao$^{1}$, and Chun-Yi Lee$^{1}$
\thanks{* indicates equal contribution}
\thanks{$^{1}$ Department of Computer Science, National Tsing Hua University}
\thanks{This work has been submitted to the IEEE for possible publication. Copyright may be transferred without notice, after which this version may no longer be accessible.}
\vspace{-1em}
}
        
\begin{document}

\maketitle
\thispagestyle{empty}
\pagestyle{empty}

\begin{abstract}
In this paper, we introduce a new concept of incorporating factorized flow maps as mid-level representations, for bridging the perception and the control modules in modular learning based robotic frameworks.  To investigate the advantages of factorized flow maps and examine their interplay with the other types of mid-level representations, we further develop a configurable framework, along with four different environments that contain both static and dynamic objects, for analyzing the impacts of factorized optical flow maps on the performance of deep reinforcement learning agents.  Based on this framework, we report our experimental results on various scenarios, and offer a set of analyses to justify our hypothesis. Finally, we validate flow factorization in real world scenarios.
\vspace{-1.15em}
\end{abstract}

\newcommand{\needfix}[1]{\textcolor{red}{#1}}
\newcommand{\needcite}[1]{\textcolor{blue}{#1}}

\section{Introduction}
\label{sec::introduction}



Combining deep neural networks (DNNs) based perception and control has long been a crucial research area for a number of robotic application domains, such as drones~\cite{sadaghi2017cad2rl, osco2021dronereview}, self-driving cars~\cite{bojarski2016e2e, sallab2017rlcar, explain-driving}, robotic arms~\cite{levine2016e2e, levine2018arms, kalashnikov2018qtopt}, etc.  To achieve this objective, a branch of researchers~\cite{bojarski2016e2e, levine2016e2e, singh2019e2e} adopted end-to-end learning frameworks, allowing robots to perform actions based on pixels of high-dimensional, unstructured RGB images. These end-to-end learning frameworks bypass
explicit visual feature learning processes and directly maps raw visual inputs to robotic actions. However, such end-to-end design lacks the flexibility of adjusting or fine-tuning, since the perception and control parts are developed as a whole. In recent years, another branch of researchers shifted their attention to modular designs, which decompose end-to-end learning frameworks to perception and control modules, allowing them to be developed separately. To bridge these modules, appropriate intermediate feature representations, which are usually referred to as \textit{mid-level representations}~\cite{midlevel-prior, chen2020mid-level, see-before-act, muller2018driving}, are used to deliver various types of information from the perception module to the control module, and form the basis of modular frameworks for many learning-based systems.

Despite the success of the naive ensemble usage of mid-level representations~\cite{midlevel-prior}, nevertheless, 
the mid-level representations considered in the prior works usually focus on spatial features extracted from a single frame, and dedicated little attention to the significance of temporal features concealed between multiple frames.  Although a few researchers~\cite{capito2020flow-driving, cv-matter} proposed to incorporate optical flow as a type of temporal feature to represent the displacements of pixels, they overlook the essential properties and characteristics of optical flows. As a result, such a type of mid-level representation still leaves room for improvement, since it is unable precisely describe the temporal information of the objects in a scene.



In light of the above motivations, the main objective of this paper is twofold.  First, we aim to introduce and investigate the concept of flow factorization, which separates a raw optical flow map into an ego flow map and an object flow map. An ego flow map is a direct result of the motion of the viewpoint, and thus provides clues about the translation and rotation of the viewpoint.  On the other hand, an object flow map is caused by the actual movements of corresponding objects between two different image frames, and thus can serve as an indicator of the speed and direction of moving objects in a scene.  Since these two types of factorized flow maps bear different physical meanings, directly leveraging raw flow maps (i.e., the flow maps directly reflecting the displacements of pixels between two image frames) as a type of mid-level representation (e.g.,~\cite{cv-matter, piergiovanni2019representation}) could deliver misleading information to the control module. We argue that both flow components are necessary and would contribute to the learning process of the control module, and thus should be separately treated. The second objective is to investigate the interplay of factorized flows with the other types of mid-level representations, and examine the impacts of different compositions of them on the control module's performance.

To achieve the above two objectives, we developed a framework for analyzing and validating the impacts of several most commonly adopted mid-level representations, including semantic segmentation, depth maps, raw optical flow maps, as well as factorized flow maps, on the performance of deep reinforcement learning (DRL) agents. The framework is built atop the Unity engine~\cite{unity-eng}, and allows DRL agents to be trained to perform collision avoidance tasks based on a configurable set of mid-level representations. To investigate the interplay of
factorized flows with the other types of mid-level representations, we designed a diversified set of evaluation scenarios with different configurations of static and dynamic objects, so as to create conditions favorable to the agents trained with different types of mid-level representations. The framework allows its dynamic objects to be configured to different speeds, such that the speed generalizability of the agents can be evaluated. The primary contributions include: 
\begin{itemize}
\vspace{-0.1em}
\item We introduce and investigate a new concept of incorporating factorized flow maps as mid-level representations.
\item We develop a configurable framework along with four different environments for analyzing the impacts of different compositions of mid-level representations.
\item We validate the complementary property of factorized flows with the other types of  mid-level representations.
\item We offer a comprehensive set of analyses for the failure cases and the impacts from changes in frame rates.
\item We validate flow factorization in real world scenarios.
\end{itemize}

\section{Background Material of DRL Based Control}
\label{sec::background}

In this section, we briefly review the background of DRL based control method, which is employed in our framework.  

\subsection{Markov Decision Process and Reinforcement Learning}
\label{subsec::mdp_and_rl}

A Markov decision process (MDP) consists of a state space $\mathbb{S}$ that contains all possible states of an environment $\mathcal{E}$, a primitive action space $\mathbb{A}$, and a reward function $\mathcal{R} : \mathbb{S} \times \mathbb{A} \to \mathbb{R}$.
In an MDP, an agent perceives a state $s_t \in \mathbb{S}$, takes an action $a_t \in \mathbb{A}$ according to its control policy $\pi: \mathbb{S}\to \mathbb{A} $, receives a reward $r_t=\mathcal{R}(x_t, a_t)$, and then transitions to a next state $s_{t+1}$ determined by $\mathcal{E}$ at each discrete timestep $t$.

The objective of reinforcement learning (RL)~\cite{intro2rl,sutton1999between} is to search for an optimal policy $\pi^*$ in $\mathcal{E}$ characterized by an MDP. An RL-based agent performs episodes of a task and iteratively updates its $\pi$ to search for $\pi^*$ via collections of transition record $(s_t, a_t, r_t, s_{t+1})$, where $\pi^*$ maximizes the expected return $ G_t=\mathbb{E}\big[ \sum^{T}_{\tau = t} \gamma^{\tau-t} r_\tau \big]$ within an episode. The discount factor $\gamma$  represents the agent's extent of preference for short-term or long-term rewards. The horizon $T$ stands for the length of one episode in $\mathcal{E}$. In recent years, RL algorithms based on DNNs, which are commonly referred to as DRL, have attracted the attention of researchers due to the fact that DRL can handle high-dimensional state spaces, e.g., the mid-level representations considered by this work~\cite{motion-perception, ficm}.

\subsection{Maximum Entropy RL and Soft Actor-Critic}
\label{subsec::SAC}



To enhance the exploration behaviors of RL agents, maximum entropy RL~\cite{energy-based-rl} proposes to find a $\pi$ that maximizes the expected return along with the entropy of $\pi$, given by:
\begin{equation}
  G_t = \mathbb{E}[\sum^T_{\tau=t}\gamma^{\tau-t}(r_\tau + \alpha\mathcal{H}(\pi))],
\end{equation}
where $\mathcal{H}(\pi)\triangleq\mathbb{E}[-\log\pi(\cdot|s_t)]$ denotes the entropy of the policy, and $\alpha$ is a temperature parameter, which controls the contribution of the entropy term to $G_t$.
Based on this concept, soft actor-critic (SAC)~\cite{sac1, sac2} further introduces a new model-free learning framework, which integrates a parameterized soft-Q function $Q_\theta$ and a parameterized stochastic policy $\pi_\phi$, where $\theta$ and $\phi$ denote the weights of DNNs. SAC achieves the state-of-the-art results on several benchmarks~\cite{openai-gym} for continuous control tasks. The weights $\theta$ and $\phi$ are trained by iteratively minimizing the objective functions $J_Q(\theta)$ and $J_\pi(\phi)$ respectively, formulated as:
\begin{equation}\label{eq:sac-value-objective}
  \resizebox{.9\hsize}{!}{$\displaystyle
  J_Q(\theta) = \mathbb{E}_{s_t,a_t}\left[
    \frac{1}{2}(Q_\theta(s_t,a_t) - (r(s_t,a_t) +\gamma \mathbb{E}_{s_{t+1}\sim p}[V(s_{t+1})]))^2
  \right],$}
\end{equation}
\begin{equation}\label{eq:sac-policy-objective}
  J_\pi(\phi) = \mathbb{E}_{s_t}[\mathbb{E}_{a_t\sim\pi_\phi}[\alpha \log(\pi_\phi(a_t|s_t))-Q_\theta(s_t,a_t)]],
\end{equation}
where $V(s_t)$ denotes the soft state value function, given by:
\begin{equation}\label{eq:sac-value-func}
  V(s_t) = \mathbb{E}_{a_t\sim\pi_\phi}[Q_\theta(s_t,a_t) - \alpha \log\pi_\phi(a_t|s_t)].
\end{equation}
While searching for a proper $\alpha$ is not straightforward,~\cite{sac2} automates this process by training $\alpha$ to minimize $J(\alpha)$:
\begin{equation}
  J(\alpha) = \mathbb{E}_{a_t\sim\pi_\phi}[-\alpha\log \pi_\phi(a_t|s_t)-\alpha\bar{\mathcal{H}}],
\end{equation}
where $\bar{\mathcal{H}}$ is the target entropy (a hyperparameter). In addition to the continuous action space setup, the authors in~\cite{discrete-sac} also extended SAC~\cite{sac2} to discrete action settings by exchanging the Gaussian distribution of the stochastic policy with a Softmax distribution. The authors in~\cite{hybrid-control} further extends SAC to a hybrid control (i.e., continuous and discrete) domain.

\section{Preliminaries of Mid-Level Representations}
\label{sec::preliminary}
In this section, we first introduce the concepts of mid-level representations, and then discuss the properties for several types of mid-level representations considered in this work. The mid-level representations discussed in this section, along with the factorized flow components introduced in Section~\ref{sec::flow_factorization}, are all incorporated into the framework presented in Section~\ref{sec::framework} for inspecting the impacts of them on the control module.

\subsection{Definition}
\label{subsec::mid-level_representation}
Mid-level representations are specific abstractions reflecting different physical or semantic meanings, and are usually bearing domain-invariant properties that are usually measured, inferred, or extracted from various kinds of visual scenes. Due to these characteristics, several types of mid-level representations, such as semantic segmentation, depth map, and raw optical flow, have long been utilized to transfer information from the perception modules to the control modules for several robotic applications~\cite{virtual-to-real, zhao2020sim2real}.
 Since different types of mid-level representations have their own strengths and weaknesses in different scenarios, a comprehensive understanding of mid-level representations, therefore, are especially crucial to the final success of modular learning-based frameworks. 


\subsection{Properties of Mid-Level Representations}
\label{subsec::properties_mid-level_representations}
In this subsection, we discuss the key properties of several commonly adopted mid-level representations, and elaborate on the advantages and primary limitations of them. 

\paragraph{Semantic segmentation ($\mathcal{S}$)} Semantic segmentation is a type of representation generated by clustering parts of an image together which belong to the same object class. It is a form of pixel-level prediction because each pixel in an image is classified according to a category. 
Semantic segmentation has received great attention by robotic and computer vision researchers over decades due to its capability to easily convey semantic meanings of images from a perception to a control modules.  Despite the convenience, its main drawback is that it is unable to reveal information about the relative positions and motions between two different objects in the spatial domain.



\paragraph{Depth map ($\mathcal{D}$)} A depth map is an image or an image channel that contains information relating to the distances of the surfaces of scene objects from a viewpoint. A depth map can be either measured by physical sensors (e.g., lidars, ultrasonic sensors, infrared sensors, etc.), or be estimated from image frames by computer vision based techniques. Depth maps are used in a wide range of robotic domains, as they reveal the spatial distances of objects in a scene. Nevertheless, depth maps are difficult to extract full semantic clues. Besides, similar to $\mathcal{S}$, they inherently do not contain the motion information of the observed objects and the viewpoint.

\paragraph{Raw optical flow map ($\mathcal{F}_{raw}$)} Optical flow estimation is a technique for evaluating the motion of objects between consecutive images, which typically requires a reference image and a target image. Optical flow is usually represented as a vector field containing displacement vectors assigned to the pixels of the reference image. These vectors indicate the shifts of the corresponding pixels from the target image, and can be exploited to represent the motion features of a scene. An optical flow map directed estimated from two image frames is referred to as a \textit{raw optical flow map ($\mathcal{F}_{raw}$)}, which contains the flow components caused by the moving objects in a scene as well as the flow components caused by the motion of the viewpoint. Except for the above information, a raw optical flow map also implicitly encodes certain extents of depth information, as the magnitudes of the optical flows from an object is related to its distance from the viewpoint. Similar to $\mathcal{D}$, $\mathcal{F}_{raw}$ also does not offer explicit semantic information, and thus is rarely used alone in modular frameworks. Moreover, it is difficult to distinguish the two different sources of the flow components concealed in $\mathcal{F}_{raw}$, and leverage them to assist the control of a robot.






\section{Optical Flow Factorization}
\label{sec::flow_factorization}

In this section, we introduce the concept of optical flow factorization, discuss the motivation of employing factorized flow components as mid-level representations, and then describe the problem formulation and the factorization procedure.

\begin{figure}[t]
  \centering
  \includegraphics[width=\linewidth]{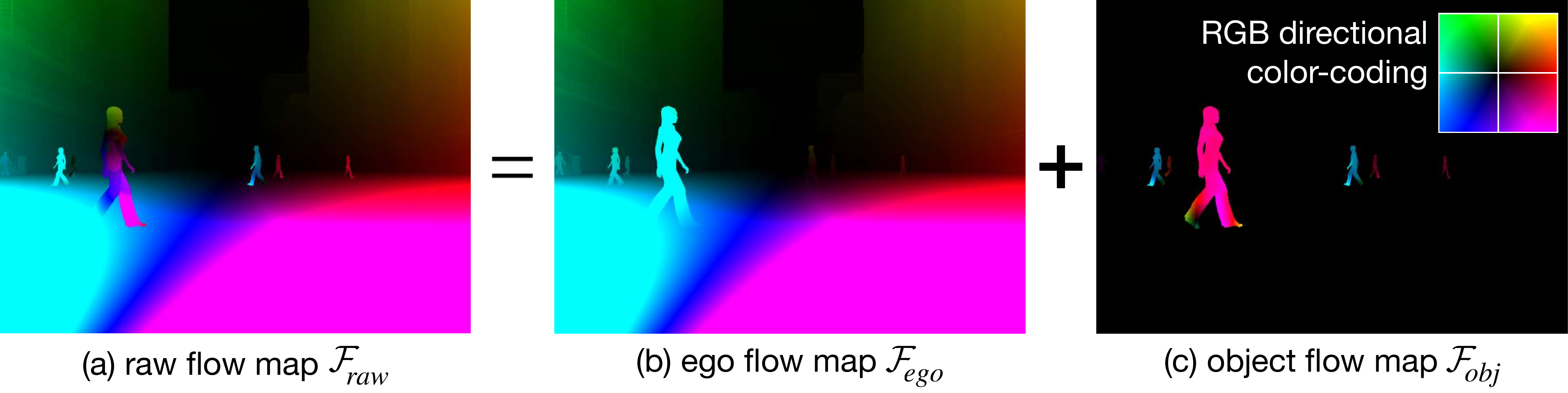}
  \caption{A visualization of flow factorization. A raw flow map ($\mathcal{F}_{raw}$) can be factorized into an ego flow map ($\mathcal{F}_{ego}$) and an object flow map ($\mathcal{F}_{obj}$). The colors presented in the flow maps are drawn based on the directional color-coding as shown at the top-right corner. The vector at each pixel coordinate corresponds to a certain flow direction and magnitude. \vspace{-1em}}
  \label{fig:flow_fact}
\end{figure}

\subsection{Concept and Motivation}

As discussed in Section~\ref{sec::introduction}, one of the primary objectives of this paper is to investigate the influences of incorporating factorized flow components as mid-level representations. In light of the limitation of $\mathcal{F}_{raw}$ discussed in Section~\ref{sec::preliminary}, we propose to further factorize it into two constituent flow fields, which are referred to as \textit{ego flow map ($\mathcal{F}_{ego}$)} and \textit{object flow map ($\mathcal{F}_{obj}$}), respectively. The former reflects the portion of $\mathcal{F}_{raw}$ caused by the motion of the viewpoint, while the latter corresponds to the rest portion of $\mathcal{F}_{raw}$ caused by the motion of the observed objects. Such a factorization allows $\mathcal{F}_{ego}$ and $\mathcal{F}_{obj}$ to explicitly represent specific physical meanings, and thus enhances the expressiveness when optical flow is intended to be utilized as a type of mid-level representation.  

Fig.~\ref{fig:flow_fact} illustrates a motivational example of optical flow factorization. In this example, the viewpoint is moving forward, while the pedestrian closest to the viewpoint is moving to the right. 
In the leftmost subfigure, the raw flow field within that pedestrian region is composed of displacement vectors pointing to multiple directions (i.e., represented in terms of different colors) due to the different moving directions of the viewpoint and the pedestrian. When the raw flow field is factorized into $\mathcal{F}_{ego}$ and $\mathcal{F}_{obj}$, as depicted on the right-hand side of Fig.~\ref{fig:flow_fact}, the flow fields within the pedestrian region become consistent in both of them.   This example implies that simply leveraging $\mathcal{F}_{raw}$ as the mid-level representation might lead to ambiguity for interpreting the motions of the objects in a scene.  On the other hand, $\mathcal{F}_{ego}$ and $\mathcal{F}_{obj}$ clearly reflects the true motions of the viewpoint and the pedestrian, and thus can serve as more expressive mid-level representations. To the best of our knowledge, the existing off-the-shelf simulators are unable to offer $\mathcal{F}_{ego}$ and $\mathcal{F}_{obj}$ as mid-level representations in a straightforward manner, if without further post processing or modification of them.


\begin{figure*}[t]
  \centering
  \includegraphics[width=\textwidth]{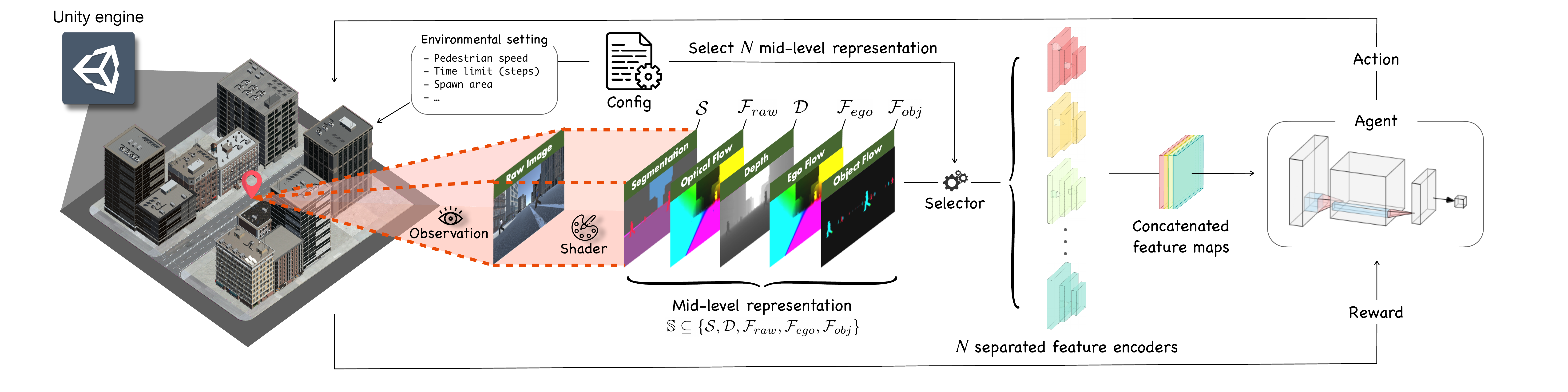}
  \caption{An overview of our framework, which allows incorporation and modification of environments, mid-level representations, scenarios, as well as DRL algorithms. Based on the given configuration, a set of mid-level representations are allowed to be selected from $\{\mathcal{S}, \mathcal{D}, \mathcal{F}_{raw}, \mathcal{F}_{ego}, \mathcal{F}_{obj}\}$, and form the state space $\mathbb{S}$ for the agent to learn a $\pi$ in a designated environment. \vspace{-1em}}
  \label{fig:overview_of_archi}
\end{figure*}

\subsection{Problem Formulation and Factorization Procedure}
\label{subsec::optical_flow_factorization}

%

\renewcommand{\vec}[1]{\mathbf{#1}}


In this subsection, we formulate the optical flow factorization problem in mathematical equations, and explain its procedure.  As defined in Section~\ref{sec::preliminary}, an optical flow map is a vector field containing displacement vectors corresponding to the pixels between two image frames.  For a certain pixel $\vec{p}_t$ of an image at timestep $t$, its corresponding raw optical flow can be denoted as a vector $\vec{f}^t_{raw} \in \mathcal{F}_{raw}$, expressed as:
\begin{equation}
\label{eq::optical-flow-factorization}
  \vec{f}^t_{raw} = \vec{f}^t_{obj} + \vec{f}^t_{ego},
\end{equation}
where $\vec{f}^t_{obj}\in \mathcal{F}_{obj}$ and $\vec{f}^t_{ego}\in \mathcal{F}_{ego}$ represent the component vectors 
caused by the motions of the object and the viewpoint, respectively. The procedure of decomposing $\vec{f}^t_{raw}$ into $\vec{f}^t_{obj}$ and $\vec{f}^t_{ego}$ at each pixel coordinate of an image is called \textit{flow factorization}. To perform the procedure, we start with the definition of $\vec{f}^t_{ego}$.
Let $\vec{p}_t=(x_t, y_t)$ be the pixel coordinate at timestep $t$, and $\vec{p}_{t'}=(x_{t'}, y_{t'})$ be the projected pixel coordinate of the same three dimensional point corresponding to $\vec{p}_t$ but
perceived from the viewpoint at a prior timestep $t'$, $\vec{f}^t_{ego}$ can be derived from the following equation:
\begin{equation}
\label{eq:pixel-displacement}
  \vec{f}^t_\text{ego} = \frac{\vec{p}_t - \vec{p}_{t'}}{\Delta t},
\end{equation}
where $\Delta t = t - t'$ denotes the time difference. Nevertheless, $\vec{p}_{t'}$ is unknown in advance, and has to be inferred from the correspondence between $\vec{p}_{t}$ and $\vec{p}_{t'}$. Given the current pixel coordinate $\vec{p}_{t}$, the rotation and translation matrices of the viewpoint $\mathbf{R}_{3\times3}$ and $\mathbf{T}_{3\times1}$, the real depth value $Z_t$, and the camera intrinsic parameters $\{f,c_x, c_y\}$, the correspondence between $\vec{p}_{t}$ and $\vec{p}_{t'}$ can be established based on the approach similar to~\cite{ego-motion-compensation}. To derive $\vec{p}_{t'}$, $\vec{p}_{t}$ is first utilized to calculate the three dimensional coordinate of the viewpoint. The pinhole camera model describes the transformation between pixel and viewpoint coordinates as a linear mapping in homogeneous coordinates, which is formulated as the following:
\begin{equation}\label{eq:pinhole-camera-model}
  \begin{bmatrix}
    x_t\\
    y_t\\
    1
  \end{bmatrix}=
  \begin{bmatrix}
    f & 0 & c_x & 0\\
    0 & f & c_y & 0\\
    0 & 0 & 1 & 0
  \end{bmatrix}
  \begin{bmatrix}
    X_t\\
    Y_t\\
    Z_t\\
    1
  \end{bmatrix},
\end{equation}
where $(X_t, Y_t, Z_t)$ is the three dimensional coordinate of the viewpoint, $f$ the focal length, and $(c_x, c_y)$ the principal point. The mappings of the two coordinates can be derived as:
\begin{equation}\label{eq:camera-to-image}
  x_t = f \frac{X_t}{Z_t} + c_x,\quad y_t = f \frac{Y_t}{Z_t} + c_y,
\end{equation}
\begin{equation}\label{eq:image-to-camera}
  X_t = \frac{Z_t}{f}(x_t-c_x),\quad Y_t = \frac{Z_t}{f}(y_t-c_y).
\end{equation}
The viewpoint coordinate of $\vec{p}_{t'}$ can be derived by applying the transform of the viewpoint using $\mathbf{R}_{3\times3}$ and $\mathbf{T}_{3\times1}$ as:
\begin{equation}\label{eq:camera-transform}
  \begin{bmatrix}
    X_{t'}\\
    Y_{t'}\\
    Z_{t'}\\
    1
  \end{bmatrix}=
  \begin{bmatrix}
    \mathbf{R}_{3\times3} & \mathbf{T}_{3\times1}\\
    \mathbf{0} & 1
  \end{bmatrix}
  \begin{bmatrix}
    X_{t}\\
    Y_{t}\\
    Z_{t}\\
    1
  \end{bmatrix}.
\end{equation}
Finally, $\vec{p}_{t'}$ can be derived from (\ref{eq:camera-to-image}), and $\vec{f}^t_\text{ego}$ can be calculated from (\ref{eq:pixel-displacement}). Typically, $\vec{f}^t_{raw}$ can be obtained from the shader of a simulation engine, or evaluated from raw RGB images by conventional flow estimation approaches. As a result, $\vec{f}^t_{obj}$ can be obtained by subtracting $\vec{f}^t_{ego}$ from $\vec{f}^t_{raw}$. By applying the above flow factorization procedure to all the pixel coordinates of an image, $\mathcal{F}_{raw}$ can be factorized into $\mathcal{F}_{ego}$ and $\mathcal{F}_{obj}$.

\begin{figure*}[t]
  \centering
  \includegraphics[width=.79\textwidth]{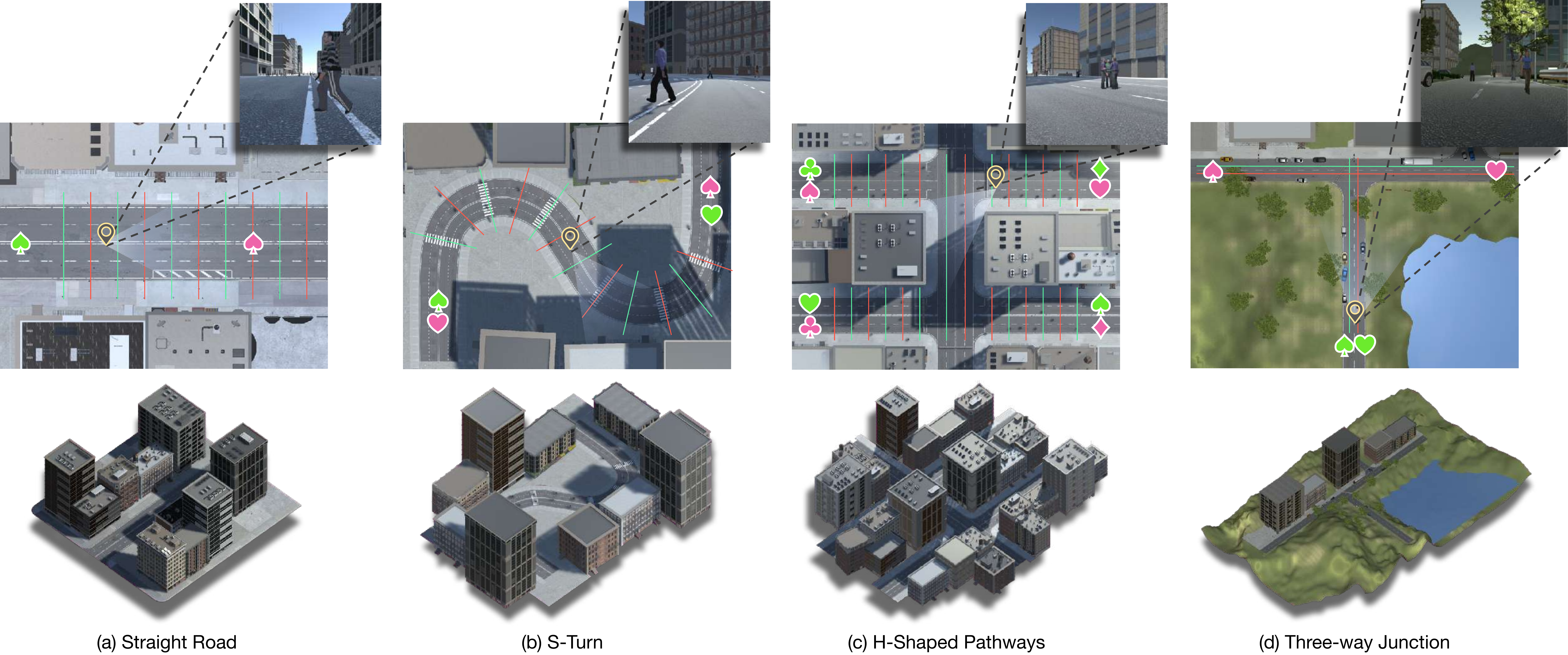}
  \caption{An overview of the environments designed for our experiments. The green and red lines represent the walking paths of the pedestrians.  The corresponding green and red suits denote the starting and ending zones of different paths, respectively. \vspace{-1em}}
  \label{fig:environment}
\end{figure*}

\section{Framework}
\label{sec::framework}

\subsection{Design Philosophy of the Framework}
\label{subsec::pilosophy}

In order to validate our assumption that $\mathcal{F}_{ego}$ and $\mathcal{F}_{obj}$ are promising candidates for mid-level representations, we develop a configurable framework based on the Unity engine~\cite{unity-eng} and the Unity ML-Agents Toolkit~\cite{ml-agents}. The framework is designed with an aim to facilitate the incorporation and modification of environments, scenarios, mid-level representations, as well as DRL algorithms. Such a design philosophy allows us to fulfill our objective of investigating the impacts of $\mathcal{F}_{ego}$ and $\mathcal{F}_{obj}$ under different conditions, and examine the interplay of them with the other types of mid-level representations. In this framework, the inputs to the DRL agents (i.e., mid-level representations) are all high-dimensional. As a result, the agents are required to interpret the provided mid-level representations, and extract necessary information concealed in them to learn its policy $\pi$.  Please note that optical flow factorization is not a built-in feature of the Unity engine, and is completely developed by ourselves.
 
\subsection{Workflow of the Framework}
\label{subsec::framework_overview}
Fig.~\ref{fig:overview_of_archi} illustrates an overview of the framework. To validate the framework, the raw RGB image observations of the environment from the viewpoint of the camera mounted on the agent are all transformed to mid-level representations by the built-in renderer.  The shader offers five different types of mid-level representations that are discussed in Sections~\ref{subsec::properties_mid-level_representations} and~\ref{sec::flow_factorization}.    For each round of simulation, an environment (i.e., $\mathcal{E}$) along with a set of mid-level representations that form the state space (i.e., $\mathbb{S} \subseteq \{\mathcal{S}, \mathcal{D}, \mathcal{F}_{raw}, \mathcal{F}_{ego}, \mathcal{F}_{obj}\}$) are selected to train an DRL agent. Different compositions of mid-level representations can be configured as $\mathbb{S}$.   At each timestep $t$, an observed state $s_t \in \mathbb{S}$ is fed into the DRL agent as its input observation. The feature embeddings of different mid-level representations are extracted by different CNNs that belong to the agent's model, and are then concatenated together for the policy of the agent to determine an action to be performed.
The environments in the framework contain multiple static objects (cars, trees, buildings, etc.) as well as dynamic objects (i.e., road crossing pedestrians), where the speeds and moving directions of the dynamic objects are fully configurable. The agent is trained with a reward function defined in the framework, with an aim to learn to successfully navigate to the designated ending zone(s) (as denoted in Fig.~\ref{fig:environment}), without colliding with any other objects (either static or dynamic) in the environment.


\section{Experimental Results}
\label{sec::experiments}


In this section, we present the experimental setups, the quantitative and qualitative results, and a set of analyses. 

\subsection{Experimental Setup}
\label{subsec::experimental_setup}

\subsubsection{Environmental Setup}
To evaluate the impacts of mid-level representations, we carefully designed four different environments based on urban scenes. These environments are crafted to reflect multiple different realistic scenarios. The scenarios which are considered in this work span over \textit{Straight Road}, \textit{S-Turn}, \textit{H-Shaped Pathways}, as well as \textit{Three-Way Junction}. The four environments are illustrated in Fig.~\ref{fig:environment},
and are described in the following paragraphs.  The detailed configurations of these environments are presented in Table~\ref{tables:env-metadata}.

\paragraph{Straight Road}
The \textit{Straight Road} environment is the simplest one-way road with several pedestrian groups crossing the road. The agent has to move on the road, avoiding the pedestrians, and not hitting any sidewalks and buildings.




\paragraph{S-Turn}
The \textit{S-Turn} environment features an s-shaped path in which the agent has to learn to keep making turns in correct directions while avoiding any obstacles in it. If there is no explicit semantic information, it will be difficult for the agent to move correctly on the road region.  As a result, using either raw optical flow map or depth map would be disadvantageous for the agent to navigate to the ending zone.


\paragraph{H-Shaped Pathways}
The \textit{H-Shaped Pathways} environment is even more difficult than the previous two, and features two intersections as well as randomly selected navigation path from four possible ones.  In addition, the environment contains static pedestrians and dynamic pedestrians.  As a result, if the agent only detects semantic information, it will be difficult for the agent to be aware of whether the pedestrians are moving or not.  The mechanism of randomly selecting the navigation path at each episode is included to test if the agent overfits.  Besides, the directions of moving pedestrians might be perpendicular or parallel to the direction of the agent.

\paragraph{Three-Way Junction}

The \textit{Three-way Junction} is designed for evaluating the agent's capability for making either left or right turns, which is considered to be more complicated than \textit{Straight Road}.  The T-shaped design aims to prevent the agent from over-fitting on any specific turn direction.  The environment contains more types of obstacles, including terrains, trees, fences, cars parked on the roadsides, traffic lights, and so on.  The directions of the moving pedestrians are designed to be parallel to the agent's path.





\begin{table}[t]
  \caption{The detailed setups of the four environments.}
  \resizebox{\linewidth}{!}{%
  \renewcommand{\arraystretch}{1.2}
  \newcommand{\mytoprule}{\toprule[1.0pt]}
  \label{tables:env-metadata}
  \centering
  \newcommand{\cmark}{\ding{51}}
  \newcommand{\xmark}{\ding{55}}
  \begin{tabular}{ c | cccc }
    \toprule
    \textbf{Environment} & Straight Road & S-Turn & H-Shaped Pathways & Three-Way Junction \\
    \hline
    Time limit $T$ (timesteps) & 200 & 250 & 300 & 250 \\
    Paths & $1$ & $2$ & $4$ & $2$ \\
    Lane width & $2$ & $1.5$ & $2$ & $1$ \\
    Object categories & $5$ & $6$ & $7$ & $11$ \\
    Turns & \xmark & \cmark & \cmark & \cmark \\
    Pedestrian (dynamic, perpendicular) & \cmark & \cmark & \cmark & \xmark \\
    Pedestrian (dynamic, parallel) & \xmark & \xmark & \cmark & \cmark \\
    Pedestrian (static) & \xmark & \xmark & \cmark & \xmark \\
    Obstacles (static) & \xmark & \xmark & \xmark & \cmark \\
    \bottomrule
  \end{tabular}}
  \vspace{-1em}
\end{table}


\begin{table}[t]
  \caption{The hyperparameters used for the SAC algorithm.} 
  \label{tables:hyperparameter}
  \centering
  \resizebox{\linewidth}{!}{%
    \renewcommand{\arraystretch}{1.2}
    \begin{tabular}{ l|l||l|l  }
      \toprule
      \textbf{Parameter} & \textbf{Value} &\textbf{Parameter} & \textbf{Value} \\
      \hline
      Optimizer & Adam & Number of hidden layers & $1$\\
      Learning rate & $3 \cdot 10^{-4}$ & Number of hidden units per layer & $512$\\
      Learning rate schedule & linear & Steps per update & $10$\\
      Initial entropy coefficient & $0.5$ & Training steps & $2 \cdot 10^{6}$\\
      Replay buffer size & $10,240$ & Target entropy & $0.2\cdot \log{\text{dim}(\mathbb{A})}$\\
      Batch size & $1,024$ & &\\
      \bottomrule 
    \end{tabular} 
  } \vspace{-0.5em}
\end{table}

\begin{figure*}[t]
  \centering
  \includegraphics[width=.9\textwidth]{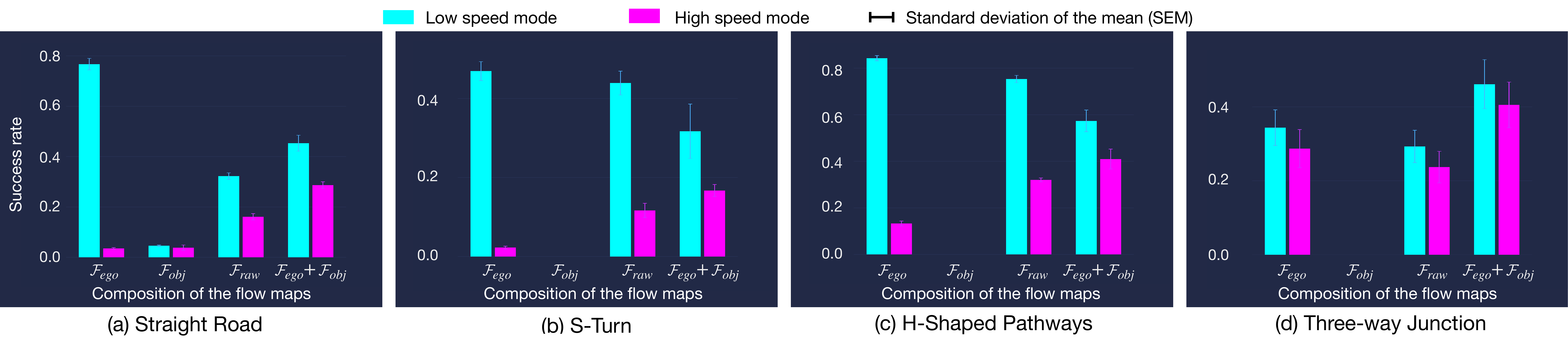}
  \caption{An impact analysis of the composition of the flow maps. The agents are trained using the normal speed mode ($1.2\sim1.8$ m/s), and evaluated under the low speed mode ($0.6\sim1.0$ m/s) and the high speed mode ($2.0\sim2.4$ m/s). \vspace{-1em}}
  \label{fig:analysis_flow_fact}
  \vspace{-0.3em}
\end{figure*}

\subsubsection{Setup of the RL Agent}
We next describe the setup of the RL agent.  At the beginning of each episode, an RL agent implemented as a DNN is trained by the SAC algorithm. The training process of the agent is completely carried out in one of the four environments (i.e., $\mathcal{E}$), and the policy $\pi$ of the agent is updated in each episode.  Within an episode, the agent interacts with $\mathcal{E}$, and receives the rewards at each timestep $t$, where the collision oracle can be obtained via the built-in physic engine.  An episode terminates when the agent reaches the ending zone, or exceeds the time limit $T$ of $\mathcal{E}$.

\paragraph{Reward Function}
In our experiments, we use the simplest reward model which only provides the minimal learning signal essential for the agents to accomplish the tasks. Specifically, the agent receives $5.0$ when reaching the ending zone, and $-5.0$ if the agent failed due to either colliding with obstacles, moving out of the bounds, or exceeding the time limit. We use $0.01$ as the survival bonus for every timestep.


\paragraph{Action Space}
The action space $\mathbb{A}$ of the agent consists of three discrete actions, and is formulated as $\mathbb{A} = \small{\{}\texttt{NO\_OP}, \texttt{TURN\_LEFT}(\alpha), \texttt{TURN\_RIGHT}(\alpha)\small{\}}$, where $\alpha$ is the angular acceleration. We use $\alpha=35\degree\per\second^2$ in the \textit{Straight Road} environment, and $\alpha=70\degree\per\second^2$ for the others. A constant forward speed $v=10\meter\per\second$ is consistently applied to the agent.

\subsubsection{Hyper-Parameter Setup} 
We employ the feature encoder of~\cite{nature-cnn} to encode each mid-level representation into an embedding. The embeddings are then concatenated and passed to the agent's value and policy networks for predictions, which are linear layers with the size equal to the action space. The detailed hyper-parameter settings are summarized in Table~\ref{tables:hyperparameter}.




\subsection{Impact Analysis of the Composition of the Flow Maps}
\label{subsec::analysis_flow_composition}

Based on the framework presented in Section~\ref{sec::framework}, we first analyze the impacts of different compositions of the flow maps, including $\mathcal{F}_{ego}$, $\mathcal{F}_{obj}$, $\mathcal{F}_{raw}$, and $\mathcal{F}_{ego}+\mathcal{F}_{obj}$, on the success rates of the DRL agents trained in the four environments discussed in the previous section, where `$\mathcal{F}_{ego}+\mathcal{F}_{obj}$' denotes that the feature embeddings of $\mathcal{F}_{ego}$ and $\mathcal{F}_{obj}$ are extracted by CNNs and then concatenated as two distinct channels.
We design three different non-overlapping speed configurations for the road-crossing pedestrians (i.e., dynamic objects): (a) low speed mode ($0.6\sim1.0$ m/s), (b) normal speed mode ($1.2\sim1.8$ m/s), and (c) high speed mode ($2.0\sim2.4$ m/s). For each mode, the pedestrians are randomly assigned a moving speed within the designated speed range. The agents are trained using the normal speed mode, and are evaluated either under the low speed mode or the high speed mode. Fig.~\ref{fig:analysis_flow_fact} depicts the evaluation results.  The observations and insights from it are discussed in the following paragraphs.


\paragraph{\textbf{$\mathcal{F}_{ego}$ is beneficial for the low speed mode, but is insufficient for the high speed mode}}
From Fig.~\ref{fig:analysis_flow_fact}, it is observed that the agents trained with only $\mathcal{F}_{ego}$ usually deliver satisfying performance when the pedestrians move slowly (i.e., the entire scene is relatively static). The rationale is that, under such circumstances, the agents could comprehend and explore the scenes by focusing only on the flows caused by the motions of the viewpoint. In other words, when the objects are all nearly static, $\mathcal{F}_{ego}$ provides the agents an easier way to capture the relative locations between it and its surrounding objects, thus allows the agent to avoid colliding with them.  For example, when the vectors in $\mathcal{F}_{ego}$ point to the left, it suggests that the viewpoint is moving or turning to the right, where the magnitudes of the vectors in $\mathcal{F}_{ego}$ reflect the relative distances of the objects to the viewpoint.
This also reveals that, by observing $\mathcal{F}_{ego}$, the agents could also learn the causal relationship between its taken action and the flow field it incurs. This, in turn, allows it to further re-adjust its next action based on the observed $\mathcal{F}_{ego}$. Although the experimental evidences suggest that the agents can benefit from $\mathcal{F}_{ego}$ in nearly static settings, however, Fig.~\ref{fig:analysis_flow_fact} also reveals that the performance of the agents drop when fast moving objects exist (i.e., in the high speed setup).  This is due to the fact that $\mathcal{F}_{ego}$ does not offer any clue about moving objects. Please note that in the \textit{Three-way Junction}, the performance of the agent does not degrade significantly in high speed mode.  This is because the moving objects in this environment only move toward or away from the agent, and thus sufficiently allows it to avoid them based on $\mathcal{F}_{ego}$.  


\paragraph{\textbf{$\mathcal{F}_{raw}$ is mostly dominated by $\mathcal{F}_{ego}$}} In \textit{S-Turn}, \textit{H-Shape Pathways}, and \textit{Three-way Junction}, it can be observed that the agents trained with $\mathcal{F}_{raw}$ perform similar to those trained with $\mathcal{F}_{ego}$ under the low speed mode. For this observation, our hypothesis is that when being fed with $\mathcal{F}_{raw}$, it could be difficult for the agents to distinguish the flows caused by the motions of the viewpoint and the moving objects. A possible explanation is that $\mathcal{F}_{obj}$ only takes up a small portion of $\mathcal{F}_{raw}$ in most cases.  As a result, $\mathcal{F}_{raw}$ is usually dominated by $\mathcal{F}_{ego}$, as depicted in Fig.~\ref{fig:flow_fact}. However, the results in Fig.~\ref{fig:analysis_flow_fact} suggest that the agents trained with $\mathcal{F}_{raw}$ still benefit from the hidden information that comes from the object motions, and achieve better performance than the agents trained with $\mathcal{F}_{ego}$ only in \textit{Straight Road}, \textit{S-Turn}, and \textit{H-Shaped Pathways} under the high speed mode.  This implies that the flows caused by the objects still plays an non-negligible role in the high speed mode. Please note that the performance under the high speed mode in the \textit{Three-way Junction} does not follow the same trend as the other environments. This is because the moving directions of the objects are mostly parallel to the moving direction of the agent.  Therefore, the contributions from  $\mathcal{F}_{obj}$ becomes relatively insignificant in this environment.



\paragraph{\textbf{$\mathcal{F}_{ego}+\mathcal{F}_{obj}$ are crucial for speed generalization}}
In can be observed from Fig.~\ref{fig:analysis_flow_fact} that the agents trained with $\mathcal{F}_{ego}+\mathcal{F}_{obj}$ achieve much balanced performance in both the low speed and the high speed modes, and are able to outperform the agents trained with the other flow compositions in all of the four environments when evaluated under the high speed mode.  This validates our hypothesis that flow factorization is crucial for interpreting the motions in a scene, and is beneficial to the scene understanding of the agents.

\begin{figure*}[t]
  \centering
  \includegraphics[width=.9\textwidth]{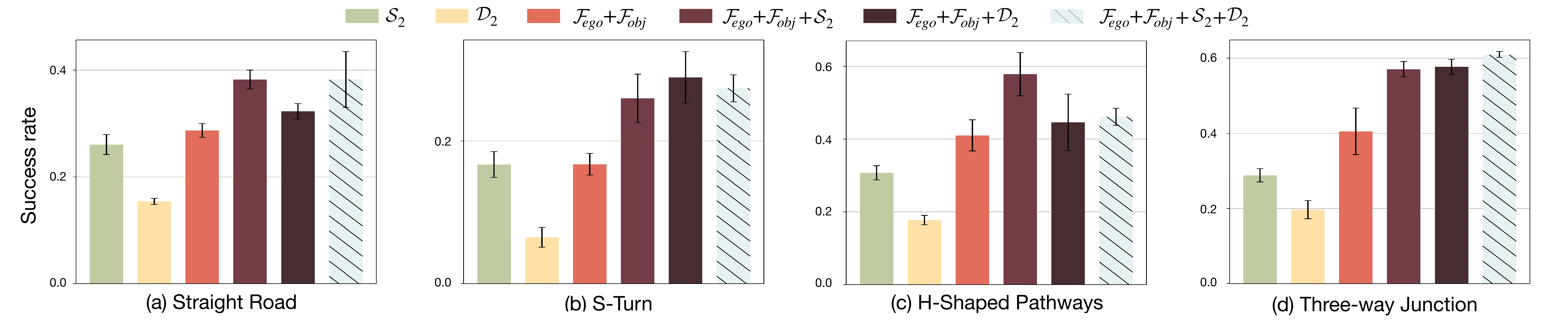}
  \caption{The experimental results for validating the complementary property  of the factorized flow maps (i.e., $\mathcal{F}_{ego}+\mathcal{F}_{obj}$). \vspace{-1em}}
  \label{fig:input_features}
\end{figure*}

\subsection{The Complementary Property of the Factorized Flows}
\label{subsec::complementary_property}

Based on the previous observations and insights, we next examine whether the factorized flow maps, i.e., $\mathcal{F}_{ego}+\mathcal{F}_{obj}$, can be used together with the other types of mid-level representations to further enhance the performance of the DRL agents.  In order to inspect if this complementary property exists, we next present an experiment for comparing the influences of the following compositions of mid-level representations on the DRL agents:  $\{\mathcal{S}_2, \mathcal{D}_2, \{\mathcal{F}_{ego}, \mathcal{F}_{obj}\}, \{\mathcal{F}_{ego}, \mathcal{F}_{obj}, \mathcal{S}_2\}, \{\mathcal{F}_{ego}, \mathcal{F}_{obj}, \mathcal{D}_2\}, \\ \{\mathcal{F}_{ego}, \mathcal{F}_{obj}, \mathcal{S}_2, \mathcal{D}_2\}\}$, where $\mathcal{S}_2$ and $\mathcal{D}_2$ denote that two consecutive semantic segmentation and depth maps are stacked, respectively.  Such a setup ensures that $\mathcal{S}_2$, $\mathcal{D}_2$, and $\{\mathcal{F}_{ego}, \mathcal{F}_{obj}\}$  are all derived on the basis of two consecutive raw RGB image frames. The above compositions are designed based on the insight from Section~\ref{subsec::analysis_flow_composition} that combined usage of $\mathcal{F}_{ego}+\mathcal{F}_{obj}$ delivers the best results in the high speed mode, and satisfying performance in the low speed mode. 


Fig.~\ref{fig:input_features} plots the results of this experiment. It can be observed that the agents trained with $\mathcal{F}_{ego}+\mathcal{F}_{obj}$ outperform those trained without them in all of the environments, suggesting the existence of the complementary property of $\mathcal{F}_{ego}+\mathcal{F}_{obj}$ with $\mathcal{S}_2$ or $\mathcal{D}_2$. Another interesting observation from Fig.~\ref{fig:input_features} is that augmenting $\mathcal{F}_{ego}+\mathcal{F}_{obj}$ with both $\mathcal{S}_2$ and $\mathcal{D}_2$ also provides positive impact on the performance of the agents. The performance boost, in our opinion, comes from the awareness of semantic information contributed by $\mathcal{S}_2$ as well as the distance information provided by $\mathcal{D}_2$. The experimental evidences therefore suggest that $\mathcal{F}_{ego}+\mathcal{F}_{obj}$ and the other mid-level representations can be utilized together, and deliver even better performance than using either of them alone.  It is also worth noting that the agents trained with $\{\mathcal{S}_2, \mathcal{D}_2, \mathcal{F}_{ego}, \mathcal{F}_{obj}\}$ do not always deliver the highest success rates.  A more complicated DNN architecture design or a more sophisticated training methodology might be necessary to enable the agents to digest and leverage the rich spatial and temporal information contained in these mid-level representations.


\subsection{Failure Case Analysis}

\begin{table}[t]
\caption{The breakdowns of the causes of the failure cases
for two different types of failures, including (1) out-of-bound (OOB) and (2) collision, in the \textit{S-Turn} environment. 
} 
\label{tables:error_analysis}
\centering
\resizebox{.9\linewidth}{!}{%
\renewcommand{\arraystretch}{1.2}
\newcommand{\mytoprule}{\toprule[1.0pt]}
\footnotesize
\begin{tabular}{ l|ccccc  }
\mytoprule
\textbf{Mid-level Representations} &
\multicolumn{3}{|c}{Causes of the Failure Cases}\\
\hline
& \textit{OOB} & \textit{Collision} & \textit{OOB-to-collision ratio} \\
\hline



$\mathcal{F}_{ego}$ + $\mathcal{F}_{obj}$ & 12.82\% & 87.18\%  & 0.147 \\

$\mathcal{D}_{2}$ & 9.51\% & 90.49\% & 0.105 \\

$\mathcal{S}_{2}$ & 4.47\% & 95.53\%  & 0.047 \\

$\mathcal{F}_{ego}$ + $\mathcal{F}_{obj}$ + $\mathcal{D}_{2}$ & 8.05\% & 91.95\%  & 0.088 \\

$\mathcal{F}_{ego}$ + $\mathcal{F}_{obj}$ + $\mathcal{S}_{2}$ & 2.72\% & 97.28\%  & 0.028 \\

$\mathcal{F}_{ego}$ + $\mathcal{F}_{obj}$ + $\mathcal{D}_{2}$ + $\mathcal{S}_{2}$ & 3.20\% & 96.80\% & 0.033 \\

\hline
\mytoprule
\end{tabular}}
\vspace{-1em}
\end{table}

To further investigate the impacts of $\mathcal{F}_{ego}+\mathcal{F}_{obj}$ as well as the other compositions of mid-level representations, we next look into the failure cases and analyze the causes of them.  Table~\ref{tables:error_analysis} reports the breakdowns of the causes of the failure cases, including (1) out-of-bound (OOB) and (2) collision. The former corresponds to the cases that the agents touch or cut into illegal regions (e.g., sidewalks), while the latter represents the cases that the agents collide with road-crossing pedestrians.
The percentages in Table~\ref{tables:error_analysis} reveal that for all the cases, collision account for the majority of the failures.  For the case of $\mathcal{F}_{raw}+ \mathcal{F}_{ego}$, the percentage of OOB is higher than that of the other cases, as flow maps are unable to provide sufficient semantic information to distinguish the difference between roads and sidewalks. The agents solely trained with $\mathcal{D}_{2}$ also face similar issues. On the other hand, the percentage of OOB is much lower for the $\mathcal{S}_{2}$ case, since semantic segmentation explicitly informs the agent about the boundaries of roads.  When semantic segmentation is consider together with factorized flow maps (i.e., $\mathcal{F}_{raw}+ \mathcal{F}_{ego}+\mathcal{S}_{2}$), the percentage of OOB is reduced by a significant margin.  The above failure analysis thus validates that different mid-level representations bear different properties, and have different influences on the behavior and performance of the agents.




\begin{figure}[t]
  \centering
  \includegraphics[width=.95\linewidth]{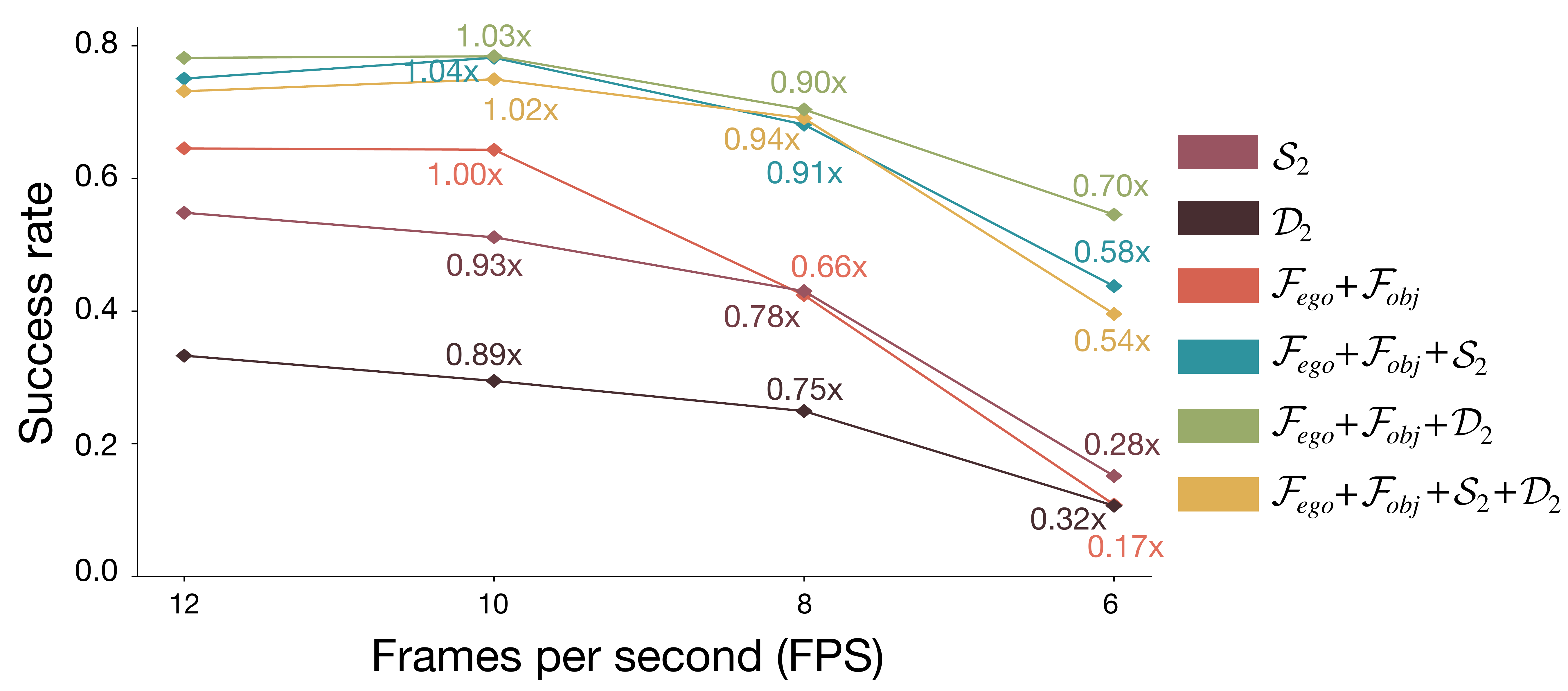}
  \caption{The performance sensitivity to the changes in FPS.}
  \label{fig:FPS}
\vspace{-1.3em}
\end{figure}


\subsection{Analysis of Sensitivity to Frame Rates}
\label{subsec::analysis_frame_rates}

As optical flow maps, $\mathcal{S}_{2}$, and $\mathcal{D}_{2}$ are obtained from two image frames, we next examine whether the time interval between these two frames would affect the performance of the agents.  Since optical flow maps reflect the displacements of pixels between two different frames, the length of the time interval between them would directly affects the magnitudes and directions of the derived flow vectors.  Any changes in this time interval would also cause $\mathcal{S}_{2}$ and $\mathcal{D}_{2}$ to be different. As a result, in this section, we further investigate on the impact of that time interval on the agents' performance under different compositions of mid-level representations.  The results are depicted in Fig.~\ref{fig:FPS}, where the time interval is reflected in the form of frames per second (FPS). The agents are trained with the frame rate set to $12$ FPS, and are evaluated using $6\sim10$ FPS under the normal speed mode.  The results in Fig.~\ref{fig:FPS} indicate that for all different compositions of mid-level representations, the performances of the agents decrease as the FPS used for evaluation decreases.  The results also show that the agents trained with $\mathcal{F}_{ego}+\mathcal{F}_{obj}$ are more sensitive to the time interval than those trained with either $\mathcal{S}_{2}$ or $\mathcal{D}_{2}$. This can be attributed to the fact that flow maps directly encode the displacements of pixels into vectors, and thus might cause the agents to suffer from larger changes in their input observations as the time interval increases.  On the other hand, the agents trained with $\mathcal{S}_{2}$ or $\mathcal{D}_{2}$ are less sensitive to the changes in the time interval, since the stacked frames might still be similar to each other and have large overlapped areas. Although the agents trained with $\mathcal{F}_{ego}+\mathcal{F}_{obj}$ suffer from larger performance drop, it still achieves similar performance to those trained with $\mathcal{S}_{2}$ or $\mathcal{D}_{2}$ when the FPS is decreased to $6$. It is also worth noting that the agents trained with 
$\{\mathcal{F}_{ego}, \mathcal{F}_{obj}, \mathcal{S}_2\}, \{\mathcal{F}_{ego}, \mathcal{F}_{obj}, \mathcal{D}_2\}$, and $\{\mathcal{F}_{ego}, \mathcal{F}_{obj}, \mathcal{S}_2, \mathcal{D}_2\}$
only degrade slightly, as the agents may benefit from the complementary property discussed above.

\begin{figure}[t]
  \centering
  \includegraphics[width=\linewidth]{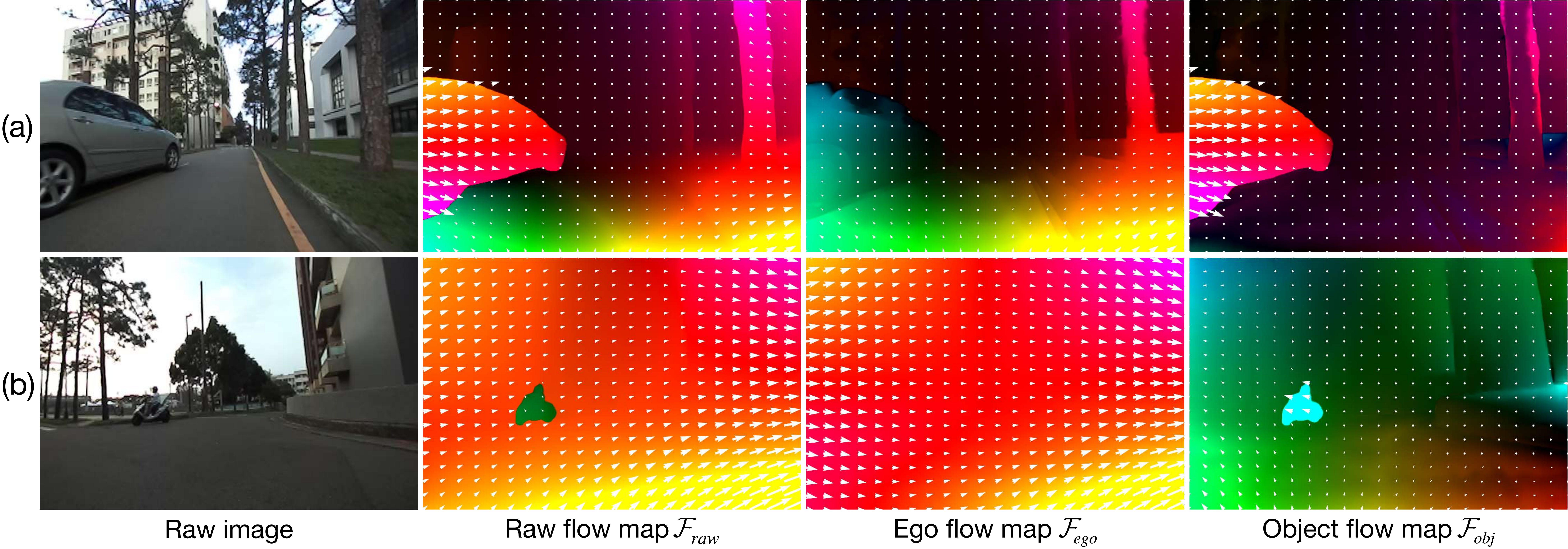}
  \caption{Real world demonstration of flow factorization.}
  \label{fig:real_flow}
  \vspace{-1.2em}
\end{figure}

\subsection{Real World Validation of Flow Factorization}
\label{subsec::real_flow_fact}

In this section, we demonstrate flow factorization in the real world scenarios.  We generate $\mathcal{F}_{raw}$ by the RAFT flow predictor \cite{teed2020raft}, obtain $Z$, $\mathbf{R}_{3\times3}$, and $\mathbf{T}_{3\times1}$ from the ZED 2i camera~\cite{zed}, and perform the flow factorization procedure discussed in Section~\ref{sec::flow_factorization}.  The results are depicted in Fig.~\ref{fig:real_flow}, in which two scenarios with different $\mathcal{F}_{raw}$ are shown in Fig.~\ref{fig:real_flow}~(a) and (b), respectively. In Fig.~\ref{fig:real_flow}~(a), it can be observed that the
boundary and the direction of the car are clearly reflected in its $\mathcal{F}_{obj}$. On the other hand, Fig.~\ref{fig:real_flow}~(b) presents a more complicated scenario, in which both the bike and the viewpoint move toward to the left, causing the magnitudes and directions of the flow vectors from $\mathcal{F}_{raw}$ to be unable to provide the correct clue about the true motion of the bike.  This, in turn, may lead to difficulties and ambiguities when performing control. With flow factorization, it can be observed from Fig.~\ref{fig:real_flow}~(b) that $\mathcal{F}_{obj}$ correctly reflects the direction of the bike (i.e., moving left). This demonstration therefore validates the applicability of the proposed concept in the real world.


\section{Conclusions}
\label{sec::conclusions}
In this paper, we introduced the use of factorized flow maps as mid-level representations. To validate the effectiveness and the benefits of combined usage of factorized flows with the other types of mid-level representations, we developed a Unity-based framework and four evaluation environments.  The experimental results suggested that factorized flow maps are complementary to the other mid-level representations, and can enhance the performance of DRL agents. We also inspected the failure cases, and showed that different mid-level representations may lead to different failure behaviors of the agents. We further examined the performance sensitivity of various configurations to FPS.  Finally, we demonstrated the applicability of flow factorization in real world scenarios.

\bibliographystyle{ieeetr}
\bibliography{reference}

\begin{thebibliography}{10}

\bibitem{sadaghi2017cad2rl}
F.~Sadeghi and S.~Levine, ``{CAD2RL:} real single-image flight without a single
  real image,'' in {\em Robotics: Science and Systems XIII}, 2017.

\bibitem{osco2021dronereview}
L.~P. Osco, J.~M. Junior, A.~P.~M. Ramos, {\em et~al.}, ``A review on deep
  learning in {UAV} remote sensing,'' {\em Int. J. Appl. Earth Obs.
  Geoinformation}, vol.~102, p.~102456, 2021.

\bibitem{bojarski2016e2e}
M.~Bojarski, D.~D. Testa, D.~Dworakowski, {\em et~al.}, ``End to end learning
  for self-driving cars,'' {\em arXiv preprint arXiv:1604.07316}, 2016.

\bibitem{sallab2017rlcar}
A.~E. Sallab {\em et~al.}, ``Deep reinforcement learning framework for
  autonomous driving,'' {\em arXiv preprint arXiv:1704.02532}, 2017.

\bibitem{explain-driving}
Éloi Zablocki, H.~Ben-Younes, P.~Pérez, and M.~Cord, ``Explainability of
  vision-based autonomous driving systems: Review and challenges,'' {\em arXiv
  preprint arXiv:2101.05307}, 2021.

\bibitem{levine2016e2e}
S.~Levine, C.~Finn, {\em et~al.}, ``End-to-end training of deep visuomotor
  policies,'' {\em J. Mach. Learn. Res.}, vol.~17, pp.~39:1--39:40, 2016.

\bibitem{levine2018arms}
S.~Levine, P.~Pastor, A.~Krizhevsky, {\em et~al.}, ``Learning hand-eye
  coordination for robotic grasping with deep learning and large-scale data
  collection,'' {\em Int. J. Robotics Res.}, vol.~37, pp.~421--436, 2018.

\bibitem{kalashnikov2018qtopt}
D.~Kalashnikov, A.~Irpan, P.~Pastor, {\em et~al.}, ``Qt-opt: Scalable deep
  reinforcement learning for vision-based robotic manipulation,'' {\em arXiv
  preprint arXiv:1806.10293}, 2018.

\bibitem{singh2019e2e}
A.~Singh, L.~Yang, C.~Finn, and S.~Levine, ``End-to-end robotic reinforcement
  learning without reward engineering,'' in {\em Robotics: Science and Systems
  XV}, 2019.

\bibitem{midlevel-prior}
A.~Sax, J.~O. Zhang, B.~Emi, {\em et~al.}, ``Learning to navigate using
  mid-level visual priors,'' in {\em CoRL 2019}, 2019.

\bibitem{chen2020mid-level}
B.~Chen {\em et~al.}, ``Robust policies via mid-level visual representations:
  An experimental study in manipulation and navigation,'' in {\em CoRL}, 2020.

\bibitem{see-before-act}
Y.~Lin, A.~Zeng, S.~Song, {\em et~al.}, ``Learning to see before learning to
  act: Visual pre-training for manipulation,'' in {\em ICRA}, 2020.

\bibitem{muller2018driving}
M.~M{\"{u}}ller, A.~Dosovitskiy, B.~Ghanem, and V.~Koltun, ``Driving policy
  transfer via modularity and abstraction,'' in {\em CoRL}, 2018.

\bibitem{capito2020flow-driving}
L.~Capito, {\"{U}}.~{\"{O}}zg{\"{u}}ner, and K.~A. Redmill, ``Optical flow
  based visual potential field for autonomous driving,'' in {\em Intelligent
  Vehicles Symposium, {IV} 2020}, pp.~885--891, {IEEE}, 2020.

\bibitem{cv-matter}
B.~Zhou, P.~Krähenbühl, and V.~Koltun, ``Does computer vision matter for
  action?,'' {\em Science Robotics}, vol.~4, no.~30, 2019.

\bibitem{piergiovanni2019representation}
A.~Piergiovanni and M.~S. Ryoo, ``Representation flow for action recognition,''
  2019.

\bibitem{unity-eng}
{\relax Unity technology}, ``Unity engine.'' \url{https://unity.com}.

\bibitem{intro2rl}
R.~S. Sutton and A.~G. Barto, {\em Reinforcement Learning: An Introduction}.
\newblock Cambridge, MA, USA: MIT Press, 1st~ed., 1998.

\bibitem{sutton1999between}
R.~S. Sutton, D.~Precup, and S.~Singh, ``Between {MDPs} and semi-{MDP}s: A
  framework for temporal abstraction in reinforcement learning,'' {\em
  Artificial Intelligence}, vol.~112, no.~1-2, pp.~181--211, 1999.

\bibitem{motion-perception}
A.~Amiranashvili, A.~Dosovitskiy, {\em et~al.}, ``Motion perception in
  reinforcement learning with dynamic objects,'' in {\em CoRL}, 2018.

\bibitem{ficm}
H.-K. Yang, P.-H. Chiang, M.-F. Hong, and C.-Y. Lee, ``Flow-based intrinsic
  curiosity module,'' in {\em IJCAI}, 2020.

\bibitem{energy-based-rl}
T.~Haarnoja, H.~Tang, P.~Abbeel, and S.~Levine, ``Reinforcement learning with
  deep energy-based policies,'' in {\em ICML}, 2017.

\bibitem{sac1}
T.~Haarnoja {\em et~al.}, ``Soft actor-critic: Off-policy maximum entropy deep
  reinforcement learning with a stochastic actor,'' in {\em ICML}, 2018.

\bibitem{sac2}
T.~Haarnoja, A.~Zhou, K.~Hartikainen, {\em et~al.}, ``Soft actor-critic
  algorithms and applications,'' {\em arXiv preprint arXiv:1812.05905}, 2019.

\bibitem{openai-gym}
G.~Brockman, V.~Cheung, L.~Pettersson, {\em et~al.}, ``Openai gym,'' {\em arXiv
  preprint arXiv:1606.01540}, 2016.

\bibitem{discrete-sac}
P.~Christodoulou, ``Soft actor-critic for discrete action settings,'' {\em
  arXiv preprint arXiv:1910.07207}, 2019.

\bibitem{hybrid-control}
O.~Delalleau {\em et~al.}, ``Discrete and continuous action representation for
  practical rl in video games,'' {\em arXiv preprint arXiv:1912.11077}, 2019.

\bibitem{virtual-to-real}
Z.~Hong, Y.~Chen, H.~Yang, {\em et~al.}, ``Virtual-to-real: Learning to control
  in visual semantic segmentation,'' in {\em IJCAI}, 2018.

\bibitem{zhao2020sim2real}
W.~Zhao, J.~P. Queralta, and T.~Westerlund, ``Sim-to-real transfer in deep
  reinforcement learning for robotics: a survey,'' in {\em SSCI}, 2020.

\bibitem{ego-motion-compensation}
R.~N. Elek, A.~I. K{\'{a}}roly, {\em et~al.}, ``Towards optical flow ego-motion
  compensation for moving object segmentation,'' in {\em ROBOVIS}, 2020.

\bibitem{ml-agents}
A.~Juliani, V.-P. Berges, E.~Teng, {\em et~al.}, ``Unity: A general platform
  for intelligent agents,'' {\em arXiv preprint arXiv:1809.02627}, 2020.

\bibitem{nature-cnn}
V.~Mnih {\em et~al.}, ``Human-level control through deep reinforcement
  learning,'' {\em Nature}, vol.~518, no.~7540, pp.~529--533, 2015.

\bibitem{teed2020raft}
Z.~Teed and J.~Deng, ``{RAFT:} recurrent all-pairs field transforms for optical
  flow,'' in {\em ECCV}, 2020.

\bibitem{zed}
Stereolabs, ``Zed stereo cameras web site.'' \url{https://www.stereolabs.com}.

\end{thebibliography}

\end{document}